%% file: main.tex
\pdfoutput=1

\documentclass[11pt]{article}

\usepackage[final]{coling}

\usepackage{times}
\usepackage{latexsym}

\usepackage[T1]{fontenc}
\usepackage[utf8]{inputenc}

\usepackage{microtype}

\usepackage{inconsolata}

\usepackage{graphicx}

\usepackage{array}
\usepackage[T1]{fontenc}
\usepackage{inconsolata}
\usepackage[utf8]{inputenc}
\usepackage{latexsym}
\usepackage{microtype}
\usepackage{times}

\usepackage[normalem]{ulem}
\useunder{\uline}{\ul}{}
\usepackage{booktabs}
\usepackage{enumitem}
\usepackage{graphicx}
\usepackage{tabularx}
\usepackage{xcolor}
\usepackage{xspace}
\usepackage{colortbl}
\usepackage{caption}
\usepackage{subcaption}
\usepackage{comment}
\usepackage{tablefootnote}
\usepackage{graphicx}
\usepackage{multirow}
\usepackage{tablefootnote}
\usepackage{hhline}
\usepackage{listings}

\newif\ifcomments
\commentstrue
\ifcomments
\newcommand{\draftcomment}[3]{{\color{#2}[\textsc{#1} #3]}}
\else
\newcommand{\draftcomment}[3]{}
\fi

\newif\ifdetailed
\ifdetailed
\newcommand{\detailed}[1]{#1}
\newcommand{\abbreviated}[1]{}
\else
\newcommand{\detailed}[1]{}
\newcommand{\abbreviated}[1]{#1}
\fi

\newcommand{\dataset}{CEHA\xspace}

\title{CEHA: A Dataset of Conflict Events in the Horn of Africa}

\author{First Author \\
  Affiliation / Address line 1 \\
  Affiliation / Address line 2 \\
  Affiliation / Address line 3 \\
  \texttt{email@domain} \\\And
  Second Author \\
  Affiliation / Address line 1 \\
  Affiliation / Address line 2 \\
  Affiliation / Address line 3 \\
  \texttt{email@domain} \\}

\author{
Rui Bai\thanks{Corresponding author.},\, 
Di Lu,\, 
Shihao Ran,\, 
Elizabeth Olson,\, 
Hemank Lamba,\, 
Aoife Cahill,
 \\
 \textbf{Joel Tetreault},\, 
  \textbf{Alex Jaimes}
\\
 Dataminr Inc.
\\
\small{
    \texttt{\{rbai,dlu,sran,elizabeth.olson,hlamba,acahill,jtetreault,ajaimes\}@dataminr.com}
}
}

\begin{document}
\maketitle
\input{0_abstract}
\input{1_introduction}
\input{2_related_work}

\input{3_dataset}
\input{4_model}
\input{5_evaluation}

\input{6_conclusion}

\section*{Ethical Considerations}
\dataset is sourced from ACLED and GDELT, and we strictly adhere to their terms of use, which permit academic usage. Since the data is collected from public sources, it does not include any personally identifiable information. We have only added Event-relevance and Event-type labels, ensuring that privacy and ethical standards are maintained.

\dataset involves human annotations from experts specialized in international development in the Horn of Africa. The annotations were conducted during the course of their professional, paid employment. 

Given the conflicting nature of events included in \dataset, we recognize the potential of \dataset being misused to spread misinformation or promote violence. To mitigate these risks, we make sure we control the access of \dataset to responsible parties and individuals by attaching a strict accessing policy and license when we release the dataset. We also urge all research utilizing \dataset to undergo ethical review and follow institutional guidelines for responsible research in this area.

\section*{Limitations}

Our dataset is constrained by several factors. Firstly, it only includes event descriptions in English, potentially missing reports written in local languages such as Amharic, Somali, and Arabic. Secondly, the dataset size is limited to 500 due to finite annotation resources and the requirement for domain expertise, restricting its usage primarily to model evaluation rather than training. Due to the limited sample size, there are fewer samples for the "No" class for event-relevance in our dataset, which differs from the actual distribution in the real world. Additionally, despite efforts to balance sampling, there are inherent imbalances in event type distributions, such as a lower number of \textsc{Climate-Related Security Risks} events, simply because they are rare. Future research could focus on expanding datasets to include local languages and exploring advanced modeling techniques such as Chain of Thought LLMs. Additionally, future work could involve extending the study to other conflict-impacted areas, thereby further enhancing the coverage of AI4SG initiatives.

\section*{Acknowledgements}
We thank our colleagues Sirene Abou-Chakra, Jessie End for their support and coordination for this project and the anonymous reviewers for their constructive comments and suggestions.

\bibliography{custom}
\input{appendix}

\end{document}

%% file: 0_abstract.tex
\begin{abstract}

Natural Language Processing (NLP) of news articles can play an important role in understanding the dynamics and causes of violent conflict. Despite the availability of datasets categorizing various conflict events, the existing labels often do not cover all of the fine-grained violent conflict event types relevant to areas like the Horn of Africa. In this paper,  we introduce a new benchmark dataset \textbf{C}onflict \textbf{E}vents in the \textbf{H}orn of \textbf{A}frica region (\dataset) and propose a new task for identifying violent conflict events using online resources with this dataset. The dataset consists of 500 English event descriptions regarding conflict events in the Horn of Africa region with fine-grained event-type definitions that emphasize the cause of the conflict. This dataset categorizes the key types of conflict risk according to specific areas required by stakeholders in the Humanitarian-Peace-Development Nexus\footnote{\url{
 https://www.un.org/peacebuilding/content/humanitarian-development-and-peace-nexus}}. Additionally, we conduct extensive experiments on two tasks supported by this dataset: Event-relevance Classification and Event-type Classification. Our baseline models demonstrate the challenging nature of these tasks and the usefulness of our dataset for model evaluations in low-resource settings with limited number of training data.

\end{abstract}

%% file: 1_introduction.tex
\section{Introduction}

Online news article resources have been pivotal for
Information Extraction~\cite{dasgupta2017crimeprofiler, singh2018natural} and Event Detection tasks~\cite{nugent2017comparison, wang2018event, hordofa2020event} when coupled with advancements in NLP over recent years.
These developments make identifying 
and summarizing events for different humanitarian and development agencies more accessible than ever~\cite{ran2023new}, 
in turn accelerating
early warning and risk mitigation, timely response and resource allocation to crisis events, and enhancing decision-making to support sustainable development~\cite{jongman2015early, lang2020earth, khatoon2021development}. 
\begin{figure}[ht]
    \centering
    \includegraphics[width=1.0\linewidth]{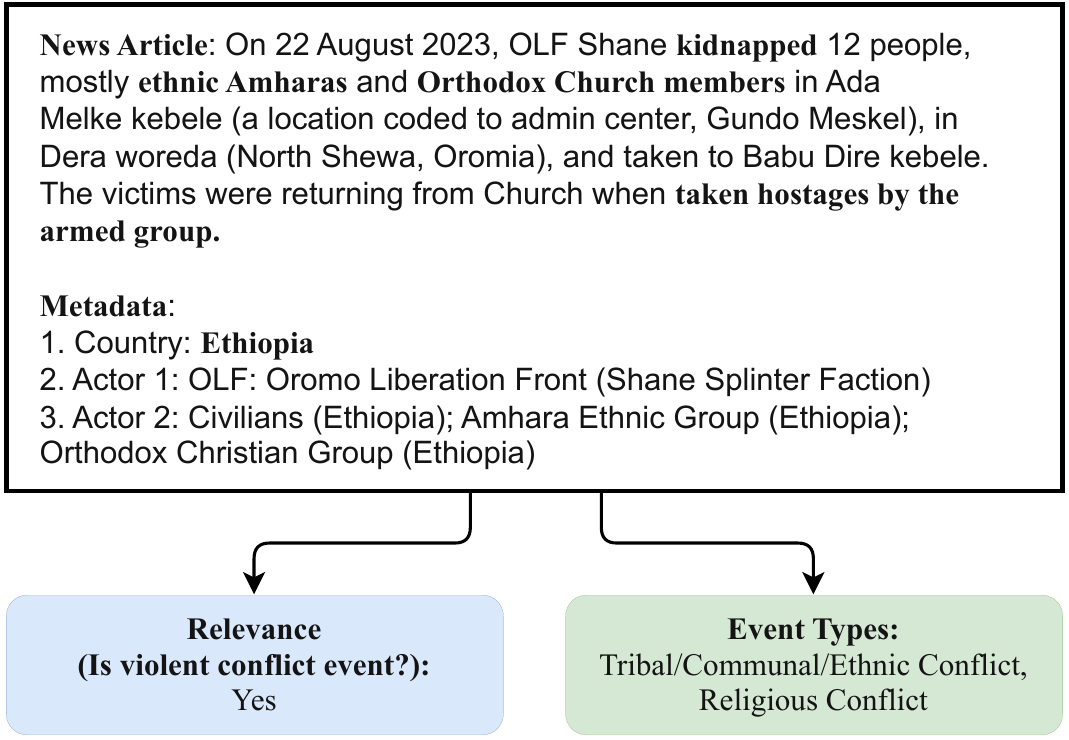}
    \caption{An example of the input/output to an NLP model for extracting event-relevance and event-types for violent conflict events in the Horn of Africa region.}
    \label{fig:undp_hook}
\end{figure}

Assessing conflict events in the regions vulnerable to crises has become increasingly crucial for humanitarian assistance.
One region in particular, the Horn of Africa\footnote{The Horn of Africa includes Djibouti, Eritrea, Ethiopia, Kenya, Somalia, Sudan, South Sudan, and Uganda.}, accounts for over 20 percent of the global caseload for humanitarian and protection assistance, with nearly 64 million people in need, according to~\citet{unocha2024}. 
Persistent conflict and volatility has shaped this urgent humanitarian crisis, including the recent armed conflict in Ethiopia's Tigray region, and ongoing civil wars in Sudan and Somalia ~\cite{kurtzer_concurrent_2022}. These conflicts stem from various complex and interconnected factors, including ethnic and religious tension, weak governance, and competition for resources~\cite{mengistu2015root, solomon2018environmental}. %

To support peacebuilding and development efforts and inform strategic interventions, %
it is essential to understand the nature and dynamics of these conflicts.
However, there are limited resources to develop NLP systems for event detection in this context. Existing event datasets like the Armed Conflict Location and Event Dataset (ACLED)~\cite{clionadh2023acled} and Global Database of Events, Language (GDELT)~\cite{leetaru2013gdelt}
only classify events based on event actions such as protests or armed clashes,  lacking systematic categorization of key event dynamics. 

To mitigate current limitations in resources for event detection in regions vulnerable to the impacts of crises like in the Horn of Africa, we propose \textbf{C}onflict \textbf{E}vents in the \textbf{H}orn of \textbf{A}frica region (\dataset),
a new dataset consisting of 500 English event descriptions from ACLED\footnote{https://acleddata.com/} and GDELT\footnote{https://www.gdeltproject.org/} covering conflict events in that region
annotated by subject matter experts.
Each event description is annotated with a binary Event-relevance label 
 to indicate if it is associated with a specific violent conflict event.  Event descriptions containing violent conflict event mentions are further annotated with 4 different event-type labels: \textsc{Tribal/Communal/Ethnic Conflict}, \textsc{Religious Conflict}, \textsc{Socio-political Violence Against Women}, and \textsc{Climate-Related Security Risks}.  Figure~\ref{fig:undp_hook} 
shows a sample event description with Event-relevance and Event-type annotations.

To summarize, our contributions are three-fold:
\begin{itemize}
\item  We publish a new benchmark dataset \dataset, containing event descriptions of violent conflicts in the Horn of Africa region to support the task for identifying and categorizing violent conflict events using online news article resources. The dataset is annotated with fine-grained event-types by subject matter experts. To the best of our knowledge, this is the first NLP dataset that pertains to this level of event regionality and event-type granularity;
\item  We conduct extensive baseline experiments for both Event-relevance and Event-type Classification with deep-learning classifiers and LLMs, demonstrating the challenging nature of this task and the usefulness of our dataset in low-resource settings with limited number of training data;

\item With \dataset, we aim to bolster the coverage of AI for Social Good (AI4SG) efforts for low-resource areas of the globe and enable more NLP research opportunities for conflict-affected %
parts of the world.
\end{itemize}

The \dataset~dataset and the code for the model training and evaluation are available at \url{https://github.com/dataminr-ai/CEHA}.

%% file: 2_related_work.tex
\section{Related Work}

\subsection{Conflict Event Datasets}

\begin{table*}[ht]
    \centering
\resizebox{\textwidth}{!}{%
\begin{tabular}{|l|l|l|l|l|l|l|}
\hline
\textbf{Dataset}    & \textbf{Focus}                              & \textbf{Geo}           & \textbf{Time}       &\textbf{Num. Events}              & \textbf{Labels}    & \textbf{Reference}                                           \\ \hline
GDELT      & Wide range of events               & Global        & 1979 - 2024 & 563 million        & Machine & \citet{leetaru2013gdelt}      \\ \hline
POLECAT    & Socio-political interactions & Global        & 2010 - 2024   & 6.2 million        & Machine & \citet{halterman2023plover}   \\ \hline
ACLED      & Political violence                 & Global        & 1997 - 2024 & 1.3 million        & Manual    & \citet{clionadh2023acled}     \\ \hline
UCDP       & Organized violence                 & Global        & 1989 - 2023    & 350,000            & Manual    & \citet{sundberg2013ucdp}      \\ \hline
GTD        & Terrorist incidents                & Global        & 1970 - 2020      & 190,000            & Manual    & \citet{lafree2007introducing} \\ \hline
SCAD       & Social conflict                    & Africa, LatAm & 1990 - 2016    & 20,000             & Manual    & \citet{salehyan2012social}    \\ \hline
HRDsAttack & Attacks on HRDs  & Global        & 2019 - 2022    & \multicolumn{1}{l|}{500} & Manual    & \citet{ran2023new}            \\ \hline
\textbf{\dataset}    & \textbf{Conflict events} & \textbf{Horn of Africa} & \textbf{2015 - 2024} & 500 & \textbf{Manual} & \textbf{Proposed dataset} \\\hline
\end{tabular}%
}
\caption{Conflict and violent event datasets.}
\label{tab:dataset_comparison}
\end{table*}

Conflict event datasets are widely developed and used by non-governmental organizations, governments, United Nations agencies, and researchers~\cite{chojnacki2012event, donnay2019integratingconflictdata, shaver2023expanding}.
These datasets have a variety of practical applications from conflict analysis and early warning, to program implementation, resource planning, and more. 

The Armed Conflict Location and Event Dataset (ACLED)~\cite{clionadh2023acled} is a seminal resource in this space, and serves as one of the data sources leveraged in the construction of \dataset. ACLED is manually curated and labelled by subject matter experts, and includes political violence and protest events sourced from traditional media, reports, online media and key informants. The Global Database of Events, Language (GDELT), also used to construct \dataset, automatically identifies and categorizes events from online print and broadcast media~\cite{leetaru2013gdelt}. In contrast to the carefully curated ACLED dataset, GDELT is much larger, with over 400 times as many different events as in ACLED and its labels are automatically generated.  %

Other key conflict event datasets include Uppsala Conflict Data Program (UCDP)~\cite{sundberg2013ucdp}, POLitical Event Classification, Attributes, and Types (POLECAT) Dataset~\cite{halterman2023plover}, Social Conflict Analysis Database (SCAD)~\cite{salehyan2012social}, and the Global Terrorism Database~\cite{lafree2007introducing}. In addition, HRDsAttack~\cite{ran2023new} presents a dataset that contains attack events around Human Rights Defenders, including various attack types such as \textsc{Killing} and \textsc{Kidnapping}, however, the geographical coverage of the dataset is global and not focused on low-resource areas. Table~\ref{tab:dataset_comparison} provides an overview of these datasets.

\subsection{Event-type Classification}
Event-type Classification is a sub-task of the Event Extraction (EE) task that aims to detect key event information such as the 5-Ws (\textit{who, what, where, when,} and \textit{why}). Most existing resources for EE such as ACE05 \cite{Doddington2004TheAC}, or its variations, such as Light ERE and Rich ERE~\cite{song2015light}, contain a wide range of event types in their event ontology, but with a limited focus on conflict event types. In the ACE ontology, 
only \textsc{Life.Injure} and \textsc{Conflict.Attack} are related to conflict events. This limited scope makes the ontology insufficient for capturing the diverse dynamics of conflict events. In HRDsAttack, the major focus  of the dataset is attack events regarding Human Rights Defenders (such as \textsc{Arbitrary Detention} or \textsc{Torture}), along with other hierarchical metadata of the event, such as \textsc{Location} and \textsc{Time}. 

In our \dataset dataset, conflict events are further categorized into four critical event-types in that region, as mentioned in reports from the Office of the Special Envoy for the Horn of Africa\footnote{https://dppa.un.org/en/mission/special-envoy-horn-of-africa}, identified by experts specializing in the Horn of Africa. These four event-types are  \textsc{Tribal/Communal/Ethnic Conflict}, \textsc{Religious Conflict}, \textsc{Socio-political Violence Against Women}, and \textsc{Climate-Related Security Risks}.

%% file: 3_dataset.tex
\section{Dataset}\label{sec:data}

\dataset is a dataset containing 500 events sourced from ACLED and GDELT, with 250 events from each source. These events were annotated by subject matter experts with experience working in International Development in the Horn of Africa region 
for event-relevance and event-type labels, utilizing well-designed annotation guidelines and various quality control measures. 
In this section, we describe how the dataset was constructed.
Section \ref{subsec:annotation_labels} introduces the iterative process of defining the annotation labels. Section \ref{subsec:datasampling} details the data sampling methods used. Finally, Section \ref{subsec:annotation_process} delves into how the annotation task was performed.

\subsection{Annotation Labels} \label{subsec:annotation_labels}

Given the complexity of categorizing conflict events ~\cite{gerner2002creation, ide2020multi}, an interdisciplinary team of experts in International Development, Crisis Risk \& Anticipation, and Computer Science collaborated to shape this project from its conception and jointly developed the annotation guidelines, event-relevance, and event-type criteria. These annotations serve as training data for models that identify and classify conflict events reported in online news sources, thereby enhancing understanding of conflict dynamics and informing strategic interventions in the Horn of Africa

While some event types have baseline definitions from ACLED's pilot projects (e.g. ACLED-Religion\footnote{https://acleddata.com/acled-religion/} in the Middle East and North Africa), which we have slightly modified, specific event types like \textsc{Tribal/Communal/Ethnic Conflict} and \textsc{Climate-Related Security Risks} are not covered. %
Definitions for these new event types were developed collaboratively with subject matter experts.

The refinement of annotation guidelines proceeded through two phases: initially, internal experts in International Development and Crisis Risk \& Anticipation refined the definitions, supplemented with positive and negative examples and detailed explanations based on an internal review involving 200 examples from these experts. %
 Subsequently, a pilot task involving 50 examples was conducted with expert annotators, whose feedback led to further definition clarity and the addition of illustrative examples.

Finally, we formalized the definitions for event-relevance and event-types, which are described in Table 
\ref{tab:relevance_def} and Table \ref{tab:eventtype_definition}. The full annotation guidelines and definitions can be found in Appendix \ref{sec:annotation_guidance}.

\begin{table*}[ht]
    \centering
    \renewcommand{\arraystretch}{1.1}
    \small
    \begin{tabularx}{\textwidth}{p{0.15\textwidth}|X}
        \multicolumn{2}{l}{\textbf{An event is defined as relevant if it meets all three criteria}}\\
        \hline
         \textbf{Criterion}&\textbf{Summarized Definition}\\
         \hhline{=|=}
         Location: Horn of Africa&The event takes place in one of the following countries: Djibouti, Eritrea, Ethiopia, Kenya, Somalia, Sudan, South Sudan, and Uganda.\\
         \hline
         Violent / Conflict Setting&The violence must be directed at a person or people rather than general expressions of anger targeting unassociated objects (e.g. burning tires or cars).\\
         \hline
        Specific Event&The text describes a specific event or incident rather than a summary of different situations.\\
  \end{tabularx}
    \caption{Event-relevance definitions (summarized). }
    \label{tab:relevance_def}
\end{table*}

\begin{table*}[ht]
    \centering
    \small
    \renewcommand{\arraystretch}{1.1}
    \begin{tabularx}{\textwidth}{p{0.15\textwidth}|p{0.43\textwidth}|X}
            \textbf{Event Type}&\textbf{Summarized Definition} & \textbf{Examples}\\
             \hhline{=|=｜=}
             Tribal/ Communal/ Ethnic Conflict& Disputes or violence involving ethnic, tribal, OR communal groups. This includes events where one or more actors had an explicit tribal, communal, clan, or ethnic affiliation received this label. & Border Guards members kidnapped a \textcolor{darkblue}{\textbf{Salamat tribal leader}} at the Hamidiya bus station in Zalingei. \textit{[The tribal leader was targeted]} \\
             \hline
             Religious Conflict& Conflicts arising from differences in religious beliefs or practices, leading to violent confrontations between religious groups. This includes events where one or more actors or targets had a stated religious affiliation (e.g. a nun, a mosque, an Islamic Militia) or individuals were targeted while engaging in religious practice (e.g. praying, visiting a mosque), even where the cause of the violence is not stated. &  \textcolor{darkblue}{\textbf{A Muslim leader}} who had denounced rebel activity and joined the army, Major Sheikh Mohammed Kiggundu, was ambushed by unidentified armed men on motorcycles. He and his escort \textcolor{darkblue}{\textbf{were killed}}. \textit{[A religious leader was attacked]}\\
             \hline
             Socio-political Violence Against Women& Civilian targeting events in which women and/or girls are the ‘target’ of the violence. This includes events where the majority of victim(s) were women and girls, and when the primary target was a woman or girl (e.g. a female politician attacked alongside her two male bodyguards) & A remote explosive \textcolor{darkblue}{\textbf{targeting a girls’ school}}. \textit{[Girls were targeted]}\\
             \hline
             Climate-Related Security Risks& Conflict events influenced by environmental and climate-related factors. Events falling into this category were required to explicitly mention both a climate related phenomenon and  a conflict event. & Clan militias ... clashed in Iarmoghe ... The area reportedly \textcolor{darkblue}{\textbf{received little rain}}, which may cause competition for pasture and explain the \textcolor{darkblue}{\textbf{clan conflict}}... \textit{[The conflict was due to lack of rain]}\\
             \hline
             Other&Events that meet the three relevancy criteria but do not fall into any of the other event types. \\
        \end{tabularx}
        \caption{Event-type definitions (summarized).}
    \label{tab:eventtype_definition}
\end{table*}

\subsection{Data Sampling} \label{subsec:datasampling}

To sample the data, we first extracted all possible violent conflict events in the Horn of Africa from both data sources and then performed balanced sampling from each.

We carefully adhered to the codebooks for each dataset to filter the data, considering the distinct structures and annotations of ACLED and GDELT. ACLED provides information about event geography, time, actors, and violent or non-violent event types labeled by specialists. It also includes summarized event descriptions. In contrast, GDELT automatically tags event information, including time, actor details and event types, following the Conflict and Mediation Event Observations (CAMEO) event coding framework~\cite{schrodt2012cameo}, which relies on keyword-based methods. %
GDELT provides links to the original articles instead of summaries. Due to ACLED's specialist-labeled data, its metadata is more trustworthy, whereas GDELT's automated tagging is less reliable.

For ACLED, we sampled event data from 2015/01/01 to 2024/01/29, focusing on events in the Horn of Africa region utilizing the \textit{COUNTRY} metadata. %
To exclude peaceful events, we filtered out events where \textit{SUB\_EVENT\_TYPE} is \textit{Agreement}, \textit{Peaceful protest} or \textit{Non-violent transfer of territory}, resulting in 97,017 events total.

Meanwhile, from the GDELT event table, we first extracted 4,390,260 events between 2020/01/01 and 2024/01/29. To filter events that happened in the Horn of Africa region, we determined the event country based on \textit{Actor1CountryCode}, \textit{Actor2CountryCode}, \textit{Actor1Geo\_CountryCode}, \textit{Actor2Geo\_CountryCode}, and \textit{ActionGeo\_CountryCode} according to the GDELT event geography ontology. %
Events were included in the dataset only if any of these fields reference a country in the Horn of Africa region. We then removed the non-violent events identified by the CAMEO Event Code in the GDELT dataset. The CAMEO ontology categorizes events into 20 groups, with the first 9 codes (01--09) representing events of cooperation between groups, and the latter 11 codes (10--20) representing conflict events between groups. Detailed codes and descriptions are provided in Appendix~\ref{sec:cameocode}. 
We specifically removed the non-violent events from groups 01 to 09. 
After filtering based on time range, geographic location, and violence level using existing labels in GDELT, we obtained 192,424 texts based on the provided URLs since GDELT does not provide the full text of the news articles. %

During the pilot annotation tasks, we noticed some data imbalance issues regarding both event-relevance and event-types: GDELT contains a significantly higher volume of irrelevant posts and a substantial number of events were annotated as \textsc{Tribal/Communal/Ethnic Conflict} in the pilot samples. To address the imbalance issue around event-relevance in GDELT before sampling the final set of articles, we applied a few-shot Mistral-Large model to remove irrelevant posts. Utilizing the annotation guidelines and examples from the pilot task as instructions to the model, it achieved an 89\% precision for the \texttt{No} class, evaluated on the second round pilot data. Detailed performance of this model and its prompt are provided in Appendix \ref{sec:prompts_data_sampling}. %
To balance the event types within the dataset, we first created targeted groups for each event type based on keyword matching on the event description and metadata provided by the original dataset (detailed criteria are listed in Table~\ref{table:sampling_criterion} in Appendix~\ref{sec:sampling_criterion}).  Next, we sampled events from each group equally for both ACLED and GDELT, with 250 events selected from each source.

\subsection{Annotation Process} \label{subsec:annotation_process}

Each data point was annotated following a two-step process: binary Event-relevance Classification, and subsequent Event-type Classification.
The annotators first determined the relevance of the event and 
for each relevant event, they then selected all relevant event type(s).

The Event-type Classification poses challenges that demand expert annotation. Annotators often rely on domain knowledge that is not explicitly stated in the text, a challenge sometimes referred to as the \textsc{Abstraction Gap}~\cite{olsen-etal-2024-socio}, e.g. that Al Shabaab is an Islamist group. Experts with expertise in the Horn of Africa annotate the \dataset.
We conducted two pilot tasks before the full task and closely monitored the full annotation process. %

\noindent
\textbf{Pilot Tasks}.
To assess the clarity and effectiveness %
of the annotation guidance and evaluate the inter-annotator agreement, we conducted 2 pilot tasks with the same 4 annotators who later performed the full task. The first pilot included 50 examples with 10 shared among annotators while the second pilot contained 20 examples, each annotated by all annotators. 
The first pilot batch revealed low agreement among annotators, prompting the refinement of annotation instructions.  This involved clarifying ambiguous cases, adding more examples, and conducting feedback sessions with annotators to enhance the accuracy of the guidelines before proceeding to the second pilot batch. As a result of these efforts, the inter-annotator agreement score for Event-relevance, measured by the average pairwise Cohen-Kappa Score, improved notably from 0.31 to 0.63. 
Table \ref{tab:pairwise_agreement} shows the detailed pairwise inter-annotator agreement score between all annotators.

\begin{table}[ht]
    \centering
    \small
    \begin{tabular}{c|cc}
        Annotator Index&Relevance&Event Type\\
        \hline
        A1&0.71&0.77\\
        A2&0.64&0.62\\
        A3&0.55&0.68\\
        A5&0.61&0.79\\
        \hline
        Average&0.63&0.72\\
    \end{tabular}
    \caption{Average pairwise Cohen Kappa score between annotators based on 20 examples in the second pilot task. Note that Annotator 4 was removed from the final annotation due to low agreement with the other annotators.}
    \label{tab:pairwise_agreement}
\end{table}

\noindent
\textbf{Full Task}.
In the full annotation task, we randomly split the data among the 4 annotators, with each annotator receiving 125 examples.
We conducted spot checks to ensure adherence to the annotation guidelines, providing feedback to annotators throughout the process.
\subsection{Data Statistics}
\dataset was randomly split into train, dev, and test sets following a 4:1:5 ratio. The annotations require expert domain knowledge, making our dataset valuable but expensive to annotate, resulting in \dataset being small though on par with other AI4SG datasets with fine-grained labels (such as \citet{ran2023new}). We used the 4:1:5 ratio to ensure a robust benchmark (test) set for evaluating models in low-resource settings for conflict events.

Table \ref{tab:data_textual_stats} presents the textual statistics of the dataset, while Table \ref{tab:data_label_stats} shows a detailed breakdown of the label statistics for \dataset. %
Event-types are only labeled for data classified as relevant events, with 9.35\% of the relevant events annotated with multiple event types and \textsc{Other} selected only when none of the 4 specified event types apply. The train, test, and dev sets are evenly distributed between ACLED and GDELT with detailed statistics listed in Table~\ref{tab:data_source_stats}.

\begin{table}[!ht]
    \centering
    \resizebox{\linewidth}{!}{
    \begin{tabular}{|l|c|c|c|c|}
    \hline
    & \textbf{train}&\textbf{dev}&\textbf{test}&\textbf{total}\\  
    \hline
    No. of articles&200&50&250&500\\
    \hline
    Total No. of tokens&32178&8565&37743&78486\\
    \hline
    Avg No. of tokens&160.89&171.30&150.97&156.97\\
    \hline
    \end{tabular}}
    \caption{Textual statistics of \dataset. }
    \label{tab:data_textual_stats}
\end{table}

\begin{table}[!ht]
    \centering
    \renewcommand{\arraystretch}{1.1}
    \resizebox{\linewidth}{!}{
    \begin{tabular}{|l|p{3.2cm}|c|c|c|c|}
    \hline
    \textbf{Task}&\textbf{Annotation Label}& \textbf{train} &	\textbf{dev} &	\textbf{test} & \textbf{total}\\ 
    \hline   
    \multirow{2}{*}{\parbox{1.5cm}{Relevance}} & Yes (relevant event) &128&32&150&310\\ \cline{2-6}
              & No (irrelevant event)&72&18&100&190\\
    \hline
    \multirow{5}{*}{\parbox{1.5cm}{Event-type}}&Tribal/ Communal/ Ethnic Conflict&51&12&52&115\\\cline{2-6}
                &Religious Conflict &41&13&28&82\\\cline{2-6}
                &Socio-political Violence Against Women&22&6&44&72\\\cline{2-6}
                &Climate-Related Security Risks&11&1&11&23\\\cline{2-6}
                &Other&14&3&30&47\\
      \hline
    \end{tabular}}
    \caption{Label statistics of \dataset.}
    \label{tab:data_label_stats}
\end{table}

\begin{table}[!ht]
    \centering
    \small
    \renewcommand{\arraystretch}{1.1}
    \begin{tabular}{|l|c|c|c|c|}
    \hline
    & \textbf{train}&\textbf{dev}&\textbf{test}&\textbf{total}\\  
    \hline
    No. of articles&200&50&250&500\\
    \hline
    ACLED&100&24&126&250\\
    \hline
    GDELT&100&26&124&250\\
    \hline
    \end{tabular}
    \caption{Source distribution of \dataset. }
    \label{tab:data_source_stats}
\end{table}

%% file: 4_model.tex
\section{Models}
In this section, we discuss the baseline models that we use to create the benchmark for the \dataset dataset. We compare two sets of models in the low-resource setting: \textbf{supervised models} (BERT~\cite{devlin2018bert}, RoBERTa~\cite{liu2019roberta}, and T5-base~\cite{raffel2023exploring})  with fine-tuning, and \textbf{prompt-based LLMs} (Mixtral 8X7B~\cite{jiang2024mixtral}, Mistral-large~\cite{mistral_2024}, DBRX~\cite{dbrx_2024}, GPT-4o~\cite{openai2024gpt4} 
and Llama3-70B~\cite{llama_3}). 

We formulate the Event-relevance Classification as a binary classification task and the Event-type Classification %
as a multi-label classification task, which is an ensemble of four binary classification tasks for each of the event types. 
We do not include the \textsc{Other} event type during training, and instead apply it to the event description when none of the four event types are assigned by the models.

\subsection{Supervised Models}
We fine-tune encoder-only models and encoder-decoder models using the  training data. For \textbf{Encoder-only Models}, we train the models using Binary Cross-Entropy Loss. For \textbf{Encoder-Decoder Models}, we use the standard maximum likelihood objective to train the model following \citet{raffel2023exploring}. 

\noindent\textbf{Encoder-only Model}
We fine-tune BERT and RoBERTa models on both classification tasks. Given the small training sample size, we only update the parameters in the last two layers. The thresholds for each class are selected based on the optimal F1 score on the dev set. 

\noindent\textbf{Encoder-decoder Model}
We select T5 %
because it is computationally efficient for fine-tuning and 
prior work~\cite{lu-etal-2023-event,ran2023new} demonstrates the effectiveness of formulating EE as Question-Answering (QA) tasks with T5 as the backbones.
For both Event-relevance and Event-type Classification, we ask T5 to answer binary questions such as \textit{Is the event relevant?}, \textit{Is the event religious conflict?} based on the context constructed from the news article content and the associated metadata. For Event-type Classification, we format the categorical ground truth label to \texttt{Yes}/\texttt{No} answer for 4 event-type question. 
The answer is \texttt{Yes} if the event type was present for this sample, otherwise \texttt{No}. 
The questions for both T5 models are listed in Table~\ref{tab:t5_questions} in Appendix~\ref{sec:t5_questions}. For Event-type Classification, we merge the model predictions to include all event types for which the model answered \texttt{Yes}.

\subsection{Prompt-based LLM Models}
We design the prompt to incorporate the annotation instructions written by our experts and propose LLM-based models for both \textbf{Zero-Shot} and \textbf{Few-Shot In-Context Learning} settings. We use Mixtral 8X7B, Mistral-large, DBRX, GPT-4o and Llama3-70B as the backbones for the experiments in Section~\ref{sec:evaluation}. All the prompts used in the experiments are detailed in Appendix~\ref{sec:llm_prompt}.

\noindent\textbf{Zero-shot Learning}
The LLMs answer directly with \texttt{Yes} or \texttt{No} to predict whether the input document is relevant or not (Event-relevance Classification Task), or predict whether the input relevant document includes a specific type of event (Event-type Classification Task). We require the model to generate the answer in the following format for easy answer parsing:
\noindent\fbox{
 \parbox{0.94\linewidth}{
<response>\\
<event\_type>Answer</event\_type>\\
\noindent<reason>reason for your selection</reason>\\
\noindent</response>
}
}
\vspace{1pt}
\noindent\textbf{Few-Shot In-Context Learning}
We implement few-shot in-context learning in chat mode, with the examples represented as parts of the conversation history. Detailed implementation of the in-context learning is listed in Appendix~\ref{apx:icl}. We use six shots for all of our experiments (three positive examples and three negative examples), based on preliminary results.  

%% file: 5_evaluation.tex
\section{Evaluation}
\label{sec:evaluation}

\subsection{Dataset and Evaluation Metrics} 
We report Precision, Recall, and F1 scores on the test set to measure the performance of various models on Event-relevance Classification and Event-type Classification\footnote{We use event-type level precision, recall and F1 score.}. The Event-type Classification is trained and evaluated on the annotated data that has been manually labeled as relevant. 

\subsection{Event-relevance Performance}
\begin{table}[!th]
\small
    \centering
    \begin{tabular}{c|ccc}
    \hline
        Models & Precision & Recall & F1 \\
        \hline
        \multicolumn{4}{c}{Supervised Models}\\
         \hline
         BERT & 63.16 &	96.00 &	76.19 \\
         RoBERTa & 72.86	& 96.67 &	83.09\\
         T5 & 78.44 & 87.33 & 82.65\\
         \hline
        \multicolumn{4}{c}{Zero-shot LLMs}\\
        \hline
         Mixtral 8X7B & 61.41 &	\textbf{98.67}	&75.70\\
         Mistral-large &67.28&	97.33	&79.56 \\
         DBRX & 71.11&	85.33&	77.58\\
         GPT-4o & 80.95&	90.67&	85.53\\
         Llama 3-70b & 72.22	&95.33	&82.18\\
         \hline
        \multicolumn{4}{c}{Few-shot In-context LLMs}\\
        \hline
         Mixtral 8X7B-6 shot & 67.61 &	96.00&	79.34\\
         Mistral-large-6 shot &78.92&	97.33&	\textbf{87.16} \\
         DBRX-6 shot & 80.12&	91.33	&85.36\\
         GPT-4o-6 shot & \textbf{88.11}&	84.00	&86.01\\
         Llama3-70b-6 shot & 87.67	&85.33&	86.49\\
    \hline
    \end{tabular}
    \caption{Performance on Event-relevance Classification Task (\%).}
    \label{tab:relevance_results}
\end{table}

Table~\ref{tab:relevance_results} shows the performance of the models on the first task. RoBERTa has the best performance among the supervised models in this low-resource setting. GPT-4o has the best performance in the zero-shot setting, and achieves comparable performance with supervised RoBERTa. All of the LLMs have better performance with Few-Shot In-Context Learning. Mistral-large and DBRX benefit more with a gain of 7.6\% and 7.78\%, respectively with In-Context Learning, and Mistral-large (six shot) achieves the best overall F1 score  (87.16\%). 

Overall, LLMs show better performance in the few-shot setting (with the only exception being Mixtral 8X7B-6 shot), which demonstrates the powerful nature of LLMs in low-resource settings due to their large amount of common world knowledge obtained via pre-training. Despite the marginal improvements in F1 scores compared to supervised models, the precision remains relatively low for most LLM model variations, which demonstrates the challenging nature of the Event-relevance Classification task. 

\subsection{Event-type Performance}
\begin{table}[!th]
\small
    \centering
    \begin{tabular}{c|ccc}
    \hline
        Models & Precision & Recall & F1 \\
        \hline
        \multicolumn{4}{c}{Supervised Models}\\
         \hline
         BERT & 52.63&	74.07&	61.54\\
        RoBERTa& 59.17&	74.07&	65.79 \\
         T5 & 79.83 & 70.37 & 74.80\\
         \hline
        \multicolumn{4}{c}{Zero-shot LLMs}\\
         \hline
         Mixtral 8x7B &  67.72	& 77.58	& 72.32\\
         Mistral-large &70.37&	80.61&	75.14 \\
         DBRX & 58.33 &	55.15	&56.70\\
         GPT-4o & 71.82	&78.79	&75.14\\
         Llama 3-70b & 71.58 &	79.39&	75.29\\
         \hline
        \multicolumn{4}{c}{Few-shot In-context LLMs}\\
         \hline
         Mixtral 8X7B-6 shot & 64.95	& \textbf{84.24}	&73.35\\
         Mistral-large-6 shot &\textbf{72.63} &	79.27	&\textbf{75.80} \\
         DBRX-6 shot & 65.46&	76.97&	70.75 \\
         GPT-4o-6 shot & 69.95&	77.58	&73.56\\
         Llama3-70b-6 shot & 67.48&	\textbf{84.24}	&74.93\\
    \hline
    \end{tabular}
    \caption{Performance on Event-type Classification Task (\%). (The scores are reported on the relevant documents.)}
    \label{tab:event_type_results}
\end{table}

\begin{table*}[!h]
\centering
\resizebox{0.98\textwidth}{!}{
\begin{tabular}{l|ccc|ccc|ccc|ccc}
\hline
Models    & \multicolumn{3}{c|}{\begin{tabular}[c]{@{}c@{}}Tribal/Communal/\\Ethnic Conflict\end{tabular}} & \multicolumn{3}{c|}{\begin{tabular}[c]{@{}c@{}}Religious\\ Conflict\end{tabular}} & \multicolumn{3}{c|}{\begin{tabular}[c]{@{}c@{}}Socio-Political Violence\\ against women\end{tabular}} & \multicolumn{3}{c}{\begin{tabular}[c]{@{}c@{}}Climate-Related\\ Security Risks\end{tabular}} \\
          & Precision & Recall & F1 & Precision & Recall & F1 & Precision & Recall & F1 & Precision & Recall & F1 \\
\hline
\multicolumn{13}{c}{Supervised Models}\\ \hline
BERT               & 56.25 & 69.23 & 62.07 & 50.00 & 75.00 & 60.00 & 58.57 & 93.18 & 71.93 & 14.29 & 18.18 & 16.00 \\
RoBERTa            & 60.71 & 65.38 & 62.96 & 52.94 & 96.43 & 68.35 & 85.00 & 77.27 & 80.95 & 22.73 & 45.45 & 30.30 \\
T5                 & 81.58 & 59.62 & 68.89 & 72.73 & 85.71 & 78.69 & 85.11 & 90.91 & 87.91 & 0.00  & 0.00  & 0.00 \\
\hline
\multicolumn{13}{c}{Zero-shot LLMs}\\ \hline
Mixtral 8X7B       & 56.63 & 90.38 & 69.63 & 75.76 & 89.29 & 81.97 & 87.80 & 81.82 & 84.71 & 100.00 & 36.36 & 53.33 \\
Mistral-large      & 61.64 & 86.54 & 72.00 & 75.76 & 89.29 & 81.97 & 82.35 & 95.45 & 88.42 & 62.50 & 45.45 & 52.63 \\
DBRX               & 71.05 & 51.92 & 60.00 & 100.00 & 46.43 & 63.41 & 100.00 & 54.55 & 70.59 & 0.00 & 0.00 & 0.00 \\
GPT-4o               & 65.22 & 86.54 & \textbf{74.38} & 70.27 & 92.86 & 80.00 & 92.31 & 81.82 & 86.75 & 80.00 & 36.36 & 50.00 \\
Llama3-70b             & 60.26 & 90.38 & 72.31 & 75.76 & 89.29 & 81.97 & 90.70 & 88.64 & \textbf{89.66} & 100.00 & 9.09  & 16.67 \\\hline
\multicolumn{13}{c}{Few-shot In-context LLMs}\\\hline
Mixtral 8X7B-6shot & 58.54 & 92.31 & 71.64 & 78.12 & 89.29 & 83.33 & 76.79 & 97.73 & 86.00 & 34.48 & 90.91 & 50.00 \\
Mistral-large-6shot& 64.18 & 82.69 & 72.27 & 80.65 & 89.29 & \textbf{84.75} & 87.50 & 81.40 & 84.34 & 63.64 & 63.64 & 63.64 \\
DBRX-6shot         & 58.90 & 82.69 & 68.80 & 87.50 & 75.00 & 80.77 & 76.92 & 90.91 & 83.33 & 35.29 & 54.55 & 42.86 \\
GPT-4o-6shot         & 65.00 & 75.00 & 69.64 & 67.57 & 89.29 & 76.92 & 88.37 & 86.36 & 87.36 & 61.54 & 72.73 & \textbf{66.67} \\
Llama3-70b-6shot       & 59.76 & 94.23 & 73.13 & 63.41 & 92.86 & 75.36 & 84.00 & 95.45 & 89.36 & 53.33 & 72.73 & 61.54 \\
\hline
\end{tabular}
}
\caption{Performance on Event-type Classification Task for each event type (\%).}
\label{tab:event_type_results_per_event}
\end{table*}

Performances for the more granular task are shown in  Table~\ref{tab:event_type_results}.  
To make a fair comparison for the Event-type Classification 
task, we evaluate the baselines on the instances marked as relevant in the expert annotation. %
T5 performs better than other supervised models  most likely since it is pretrained on a wide range of NLP tasks, it can deal with extremely low-resource settings better than the other two supervised models.

Similarly, the best zero-shot LLM (Llama3 with an F1 score of $75.29\%$) has comparable performance with the best-performing supervised model (T5 with an F1 score of $74.80\%$). However, in-context examples do not consistently provide improvement. GPT-4o and Llama3 have a slight performance drop in the six-shot setting. Mistral-large in a six-shot setting achieves the best F1 score. And DBRX benefits the most with In-Context Learning and obtains a gain of 14.05\% in F1 score.

The performance metrics for each event type from all models are detailed in  
 Table~\ref{tab:event_type_results_per_event}.
At a high level, we see a similar trend of model performance for \textsc{Tribal/Communal/Ethnic Conflict}, \textsc{Religious Conflict}, and \textsc{Socio-political Violence Against Women} -- the LLMs generally perform better than the supervised models by small margins. GPT4 has the highest F1 score (74.38\%) on the \textsc{Tribal/Communal/Ethnic Conflict} event type, with a 5.69\% increase over the best-performing supervised model T5. For \textsc{Religious Conflict}, Mistral-large-6shot achieves the best F1 score of 84.75\%, 6.06\% better than T5. The performance difference gets smaller across models for the \textsc{Socio-political Violence Against Women} event type with the highest performance coming from Llama3. Perhaps unsurprisingly, for the \textsc{Climate-Related Security Risks} event type, supervised models struggle due to the limited number of samples in the training data for this event type, with T5 failing to generate any predictions for this event type. On the other hand, LLMs understandably  stand out in this extremely low-resource setting -- most LLM models achieve much better performance for this event type, with the exception of DBRX and Llama3.

From Table~\ref{tab:data_label_stats} and Table~\ref{tab:event_type_results_per_event}, one can see that the performance of the supervised models does not scale with the number of available training samples for each event type. Similar trends can be observed for the LLM counterparts. The variations in F1 scores can be viewed as an indicator of the task difficulty for each event type: \textsc{Climate-Related Security Risks} being the most challenging event type and \textsc{Socio-political Violence Against Women} being the easiest event type to classify out of all four event types. This observation can be further backed up by the broader view of how much existing resources for different aspects of world events are available and used for model pre-training, for both supervised models and LLMs. Our hypothesis and assumption is that existing NLP resources focus more on socio-political events and less on climate-related events in low-resource areas of the globe, which is then reflected in our task and benchmark scores. This is also why we advocate for more AI4SG opportunities for low-resource and crisis-prone parts of the world given the gaps in existing resources and downstream model performance. We noticed that the LLMs have much higher recall on \textsc{Tribal/Communal/Ethnic Conflict} (94.23\% from the best prompt-based LLM), compared with supervised models (69.23\% from the best supervised model). It indicates that the common sense knowledge embedded in the LLMs is not efficient enough to identify those events. For example, the Mistral-large model mistakenly classifies the event \textit{`On 11 August 2021, members of TPLF forces raped a 60-year-old woman (Amhara) in Kebele 04 in Weldiya town (North Wello, Amhara).'} as a \textsc{Tribal/Communal/Ethnic Conflict} event as opposed to \textsc{Socio-political Violence Against Women}, because the identified actor, TPLF, is a commonly known ethnic group.

%% file: 6_conclusion.tex
\section{Conclusions}
In this paper, we present \dataset, a new dataset that aims to bridge the gap in existing NLP resources for regions vulnerable to violence, 
as in the Horn of Africa.  
Following carefully crafted annotation guidelines and quality control measures, \dataset contains 500 English online news articles annotated by subject matter experts in the field for the tasks of conflict Event-relevance Classification and fine-grained Event-type Classification. In addition, we conduct extensive experiments to demonstrate the usefulness of our dataset and the challenging nature of the new task in low-resource settings. With \dataset, we hope to inspire more NLP research interest into violent conflict event detection in %
conflict-affected regions, and to aid AI4SG efforts in general. %

%% file: appendix.tex
\appendix

\section{Annotation Guidance}
\label{sec:annotation_guidance}

\begin{figure*}
     \centering
    \includegraphics[width=\textwidth]{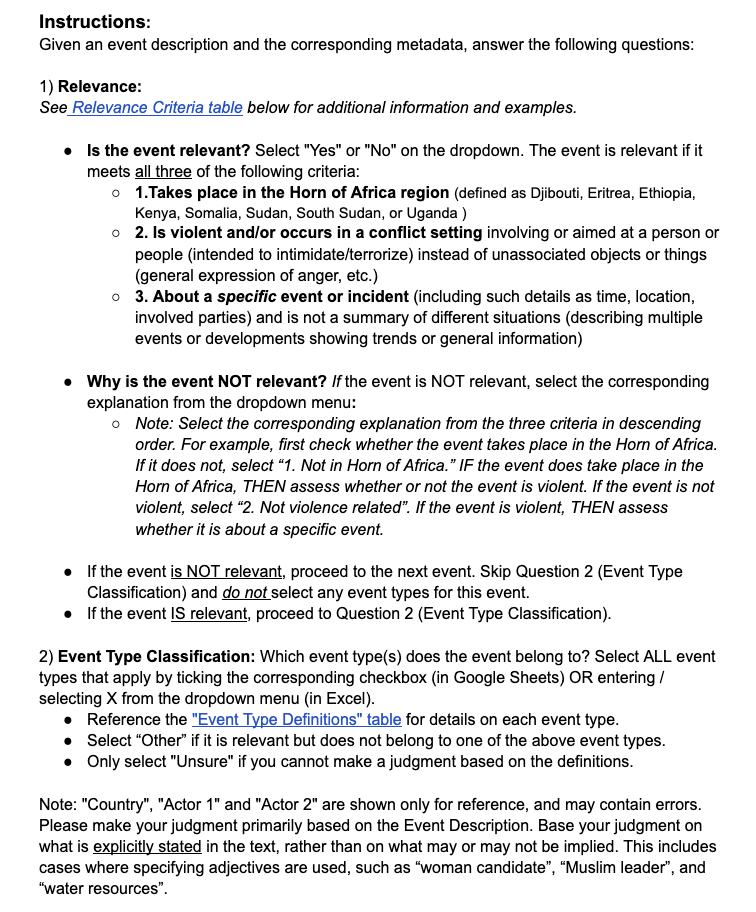}
    \caption{Screenshot of the annotation guidance (1/5).}
    \label{fig:annotation_guidance_1}   
\end{figure*}

\begin{figure*}
     \centering
    \includegraphics[width=\textwidth]{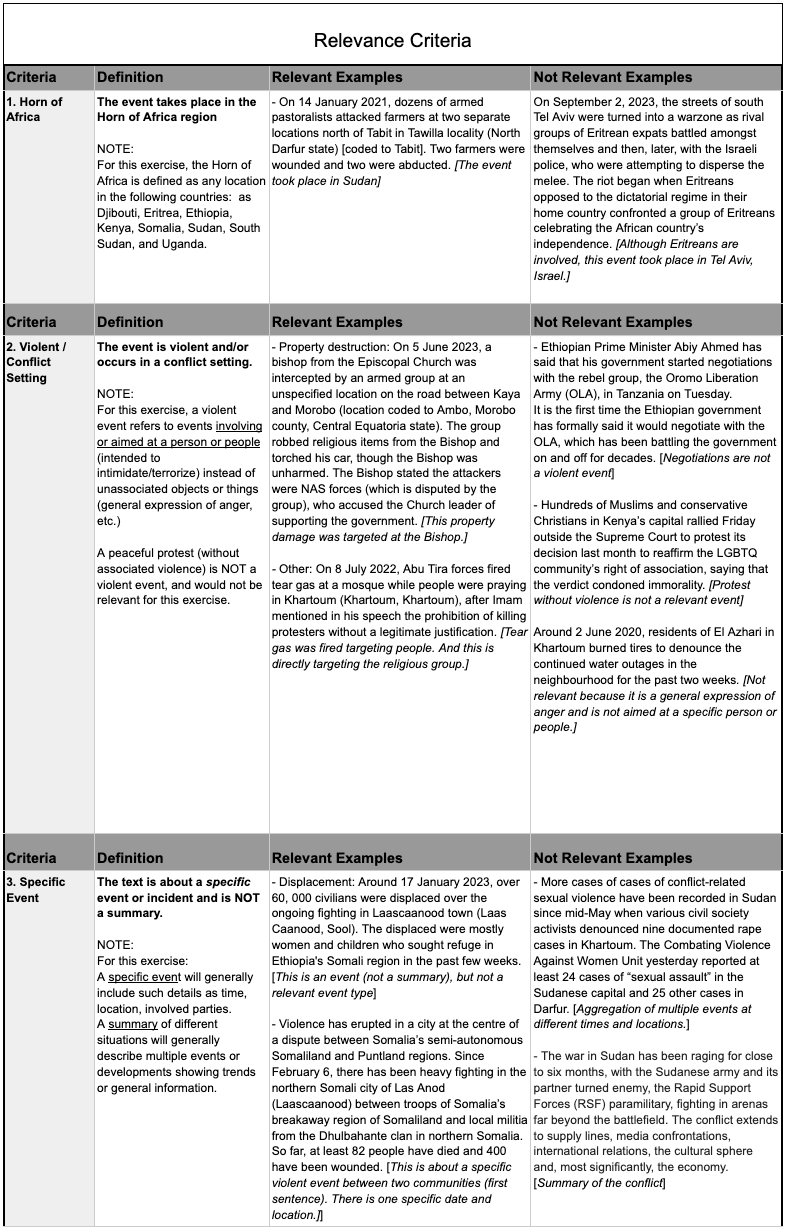}
    \caption{Screenshot of the annotation guidance (2/5).}
    \label{fig:annotation_guidance_2}   
\end{figure*}

\begin{figure*}
     \centering
    \includegraphics[width=\textwidth]{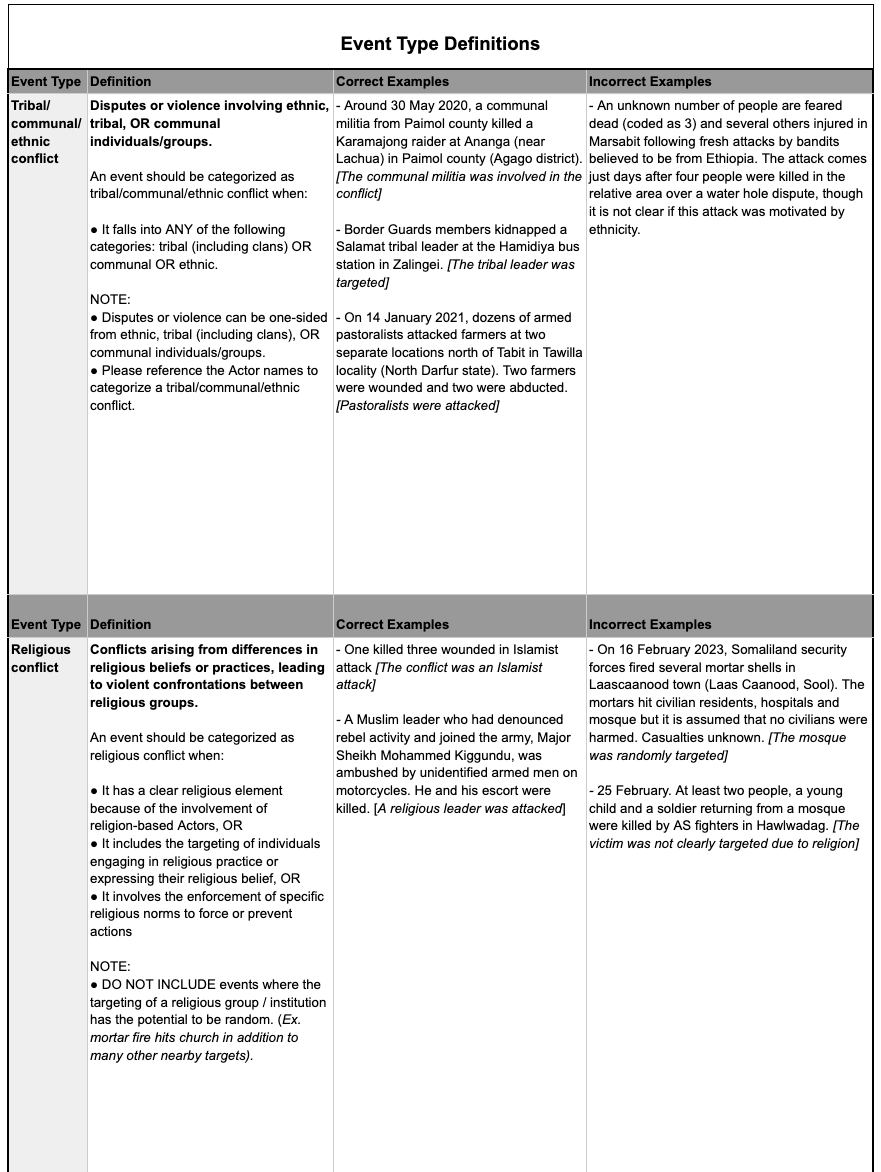}
    \caption{Screenshot ofthe annotation guidance (3/5).}
    \label{fig:annotation_guidance_3}   
\end{figure*}

\begin{figure*}
     \centering
    \includegraphics[width=\textwidth]{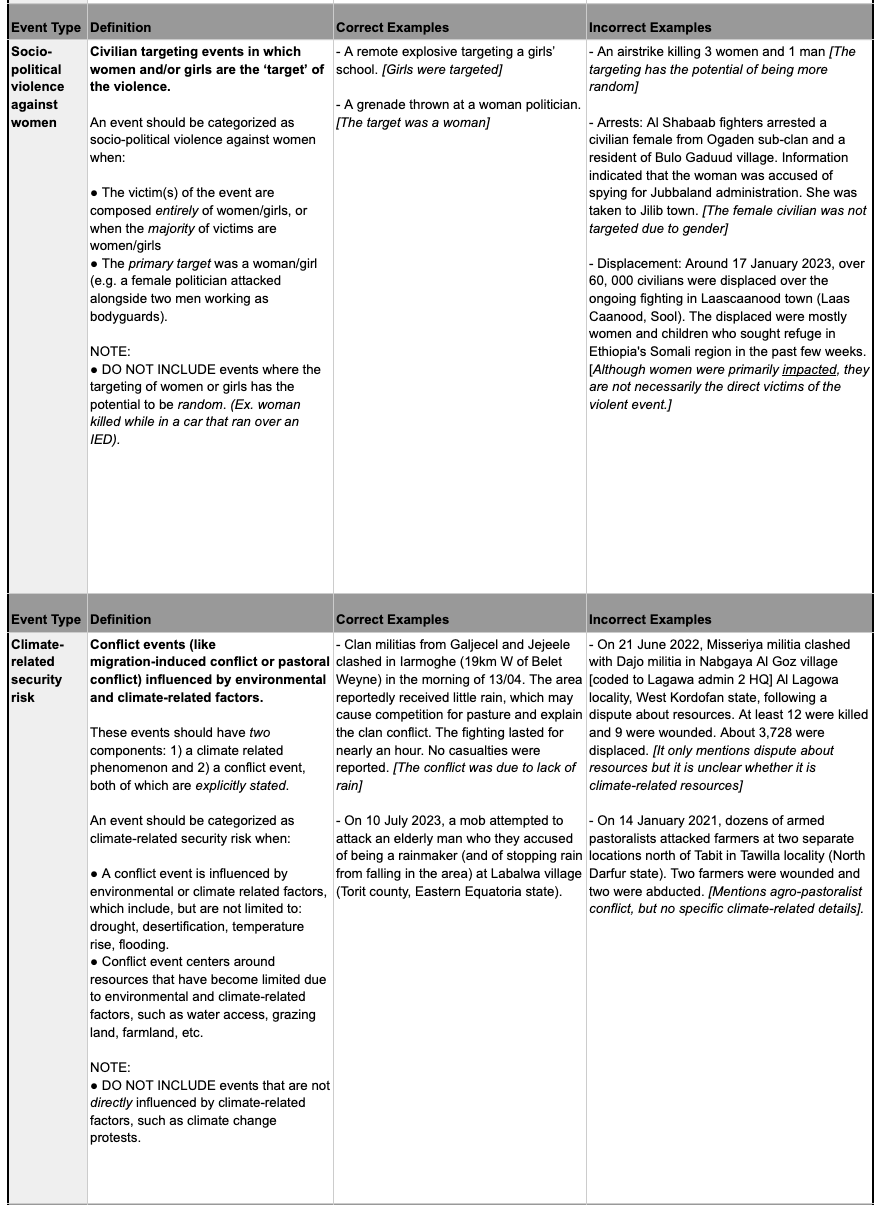}
    \caption{Screenshot of the annotation guidance (4/5).}
    \label{fig:annotation_guidance_4}   
\end{figure*}

\begin{figure*}
     \centering
    \includegraphics[width=\textwidth]{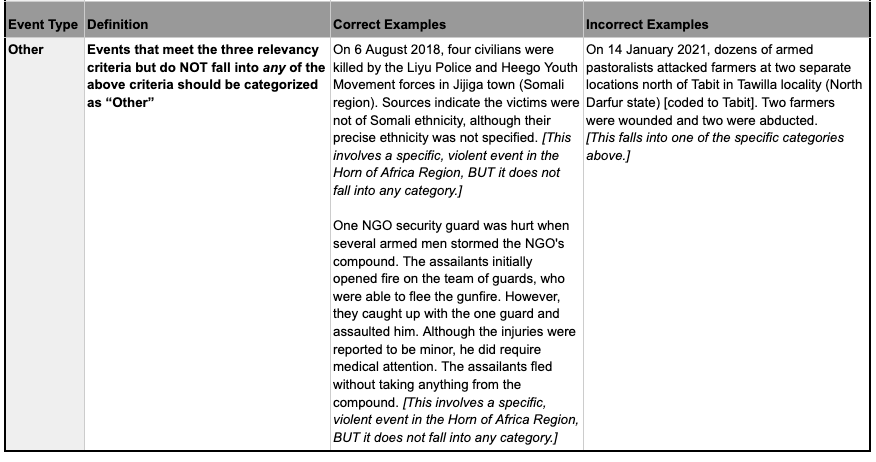}
    \caption{Screenshot of the annotation guidance (5/5).}
    \label{fig:annotation_guidance_5}   
\end{figure*}

\begin{figure*}
     \centering
    \includegraphics[width=\textwidth]{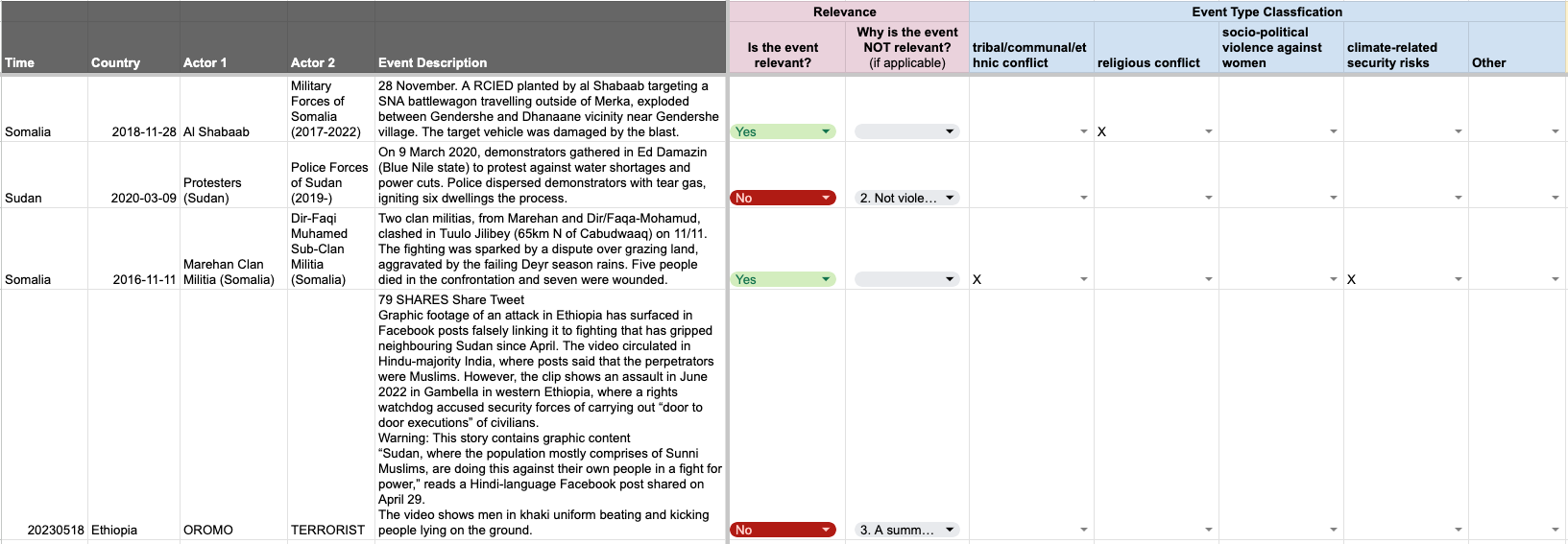}
    \caption{Screenshot of the annotation interface.}
    \label{fig:annotation_interface}   
\end{figure*}

Figures \ref{fig:annotation_guidance_1} to \ref{fig:annotation_guidance_5} show the detailed annotation guidance which is shared with all annotators and Figure \ref{fig:annotation_interface} is a screenshot of the annotation interface.

\section{CAMEO Event Code in GDELT dataset}
\label{sec:cameocode}
Table ~\ref{tab:cameocode} shows desciptions of the CAMEO Event Code used in GDELT dataset to classify events. Based on these definitions, we removed events with codes 01–09, which have a higher likelihood of being non-violent.

\begin{table}[ht]
    \centering
    \renewcommand{\arraystretch}{1.1}
    \resizebox{0.48\textwidth}{!}{
    \begin{tabular}{|c|c|}
    \hline
        \textbf{CAMEO Event Code}&\textbf{Description}\\
        \hline
        01&MAKE PUBLIC STATEMENT\\
        02&APPEAL\\
        03&EXPRESS INTENT TO COOPERATE\\
        04&CONSULT\\
        05&ENGAGE IN DIPLOMATIC COOPERATION\\
        06&ENGAGE IN MATERIAL COOPERATION\\
        07&PROVIDE AID\\
        08&YIELD\\
        09&INVESTIGATE\\
        10&DEMAND\\
        11&DISAPPROVE\\
        12&REJECT\\
        13&THREATEN\\
        14&PROTEST\\
        15&EXHIBIT FORCE POSTURE\\
        16&REDUCE RELATIONS\\
        17&COERCE\\
        18&ASSAULT\\
        19&FIGHT\\
        20&USE UNCONVENTIONAL MASS VIOLENCE\\
        \hline
    \end{tabular}}
    \caption{CAMEO code descriptions.}
    \label{tab:cameocode}
\end{table}

\section{Mistral-large Model for Data Sampling}
\label{sec:prompts_data_sampling}
The model used for relevance filtering is based on Mistral-large with few-shot in-context learning. The model is evaluated on our second round pilot experiment, achieving precision, recall, and F1 of 89\%, 38\%, and 53\%, respectively on the \texttt{No} class. %
The prompt for this model is given below. Some examples in the prompt are skipped for brevity, but the full prompt can be found in the distributed code repository.%

\noindent\small\fbox{
 \parbox{\linewidth}{
        f"""
Instruction: You are a state-of-the-art event detection system. Given a news article regarding a specific event, your job is to classify if the article is relevant based on a given set of guidelines. The article is relevant if:
1. the event it describes takes place in the Horn of Africa Region, which includes Djibouti, Eritrea, Ethiopia, Kenya, Somalia, Sudan, South Sudan, or Uganda.
2. the event it describes is violent and/or occurs in a conflict setting involving or aimed at a person or people (intended to intimidate/terrorize) instead of unassociated objects or things (general expression of anger, etc.).
3. the article describes a *specific event* and is not a summary of multiple events or different events, i.e., it is not describing multiple events or developments showing trends or general information. If an article mentions more than ONE event, it is not relevant in our setting.

Here are some examples:

The following articles are NOT violence related given the above guidelines:
1. Ethiopian Prime Minister Abiy Ahmed has said that his government started negotiations with the rebel group, the Oromo Liberation Army (OLA), in Tanzania on Tuesday. [Negotiations are not a violent event];

...

The following articles ARE specific events, NOT a summary:
1. Foreign Affairs Cabinet Secretary Alfred Mutua now says that the move to deploy Kenyan police to Haiti is not only about peace and security. In a statement shortly after the United Nations Security Council voted to allow Kenyan troops into the Caribbean country, Mutua said that it is also about rebuilding Haiti. [A UN vote regarding troop deployment is not a violent event];

...

Given the guidelines and examples above, you should answer only Yes if the article below is relevant based on the guidelines, No if the article is not relevant, Unsure if you cannot make the judgment based on the provided information. Followed by a concise description of the reason. Do not be conversational.

\{document\}. This event was POSSIBLY reported in \{country\}.

Is this article relevant based on the guidelines: 
"""
}
}

\begin{table*}[!ht]
    \small
    \renewcommand{\arraystretch}{1.1}
    \begin{subtable}{1\linewidth}
    
        \begin{tabular}{ |p{3cm}| p{7cm} | p{5cm} |}
        \hline
            \textbf{Event Type}&\textbf{Text Keywords}&\textbf{Additional Criterion}\\
            \hhline{|=|=|=|}
            Tribal/Communal/Ethnic Conflict&ethnic, communal, tribal, clan&Actor Info mentions ethnic, communal, tribal or clan\\
             \hline
            Religious Conflict&muslim, christian, mosque, church, religious, religion, islam&Actor Info mentions muslim or christian\\
             \hline
            Socio-political Violence Against Women&women, woman, girl, girls, female, gender&SUB\_EVENT\_TYPE as "Sexual violence"; Associate Actor as "Women(country)"; Contain Tag for "women targeted"\\
             \hline
            Climate-Related Security Risks&water shortage, water outage, water scarcity, water resource, resource (excluding human resource), climate, rain, rainy, flood, flooding, desert, drought, environment, environmental&None\\ 
            \hline
        \end{tabular}
        \caption{ACLED}
        
        \begin{tabular}{ |p{3cm}| p{7cm} | p{5cm} |}
        
        \hline
            \textbf{Event Type}&\textbf{Text Keywords}&\textbf{Additional Criterion}\\
            \hhline{|=|=|=|}
            Tribal/Communal/Ethnic Conflict&ethnic, communal, tribal, clan&Actor Ethnic Info is provided\\
             \hline
            Religious Conflict&muslim, christian, mosque, church, religious, religion, islam&Actor Religion Info is provided\\
             \hline
            Socio-political Violence Against Women&women, woman, girl, girls, female, gender&Event Code as "Sexually assault”\\
             \hline
            Climate-Related Security Risks&water shortage, water outage, water scarcity, water resource, resource (excluding human resource), climate, rain, rainy, flood, flooding, desert, drought, environment, environmental&None\\ 
            \hline
        \end{tabular}
        \caption{GDELT}
    \end{subtable}
    \caption{Criteria to create targeted group for each event type.}
    \label{table:sampling_criterion}
\end{table*}
\section{Balanced Data Sampling Criteria for Event Type}\label{sec:sampling_criterion}

To balance the event types for both ACLED and GDELT in \dataset, we initially create specific groups for each event type. The groups are created by employing keyword matching on the event description and combinations of the metadata from the original datasets. Table \ref{table:sampling_criterion} are the detailed criteria we utilized for both datasets. 

\section{Prompts}
\label{sec:llm_prompt}
\subsection{Implementation of In-context Learning}
\begin{figure}[!ht]
    \centering
    \includegraphics[width=\linewidth]{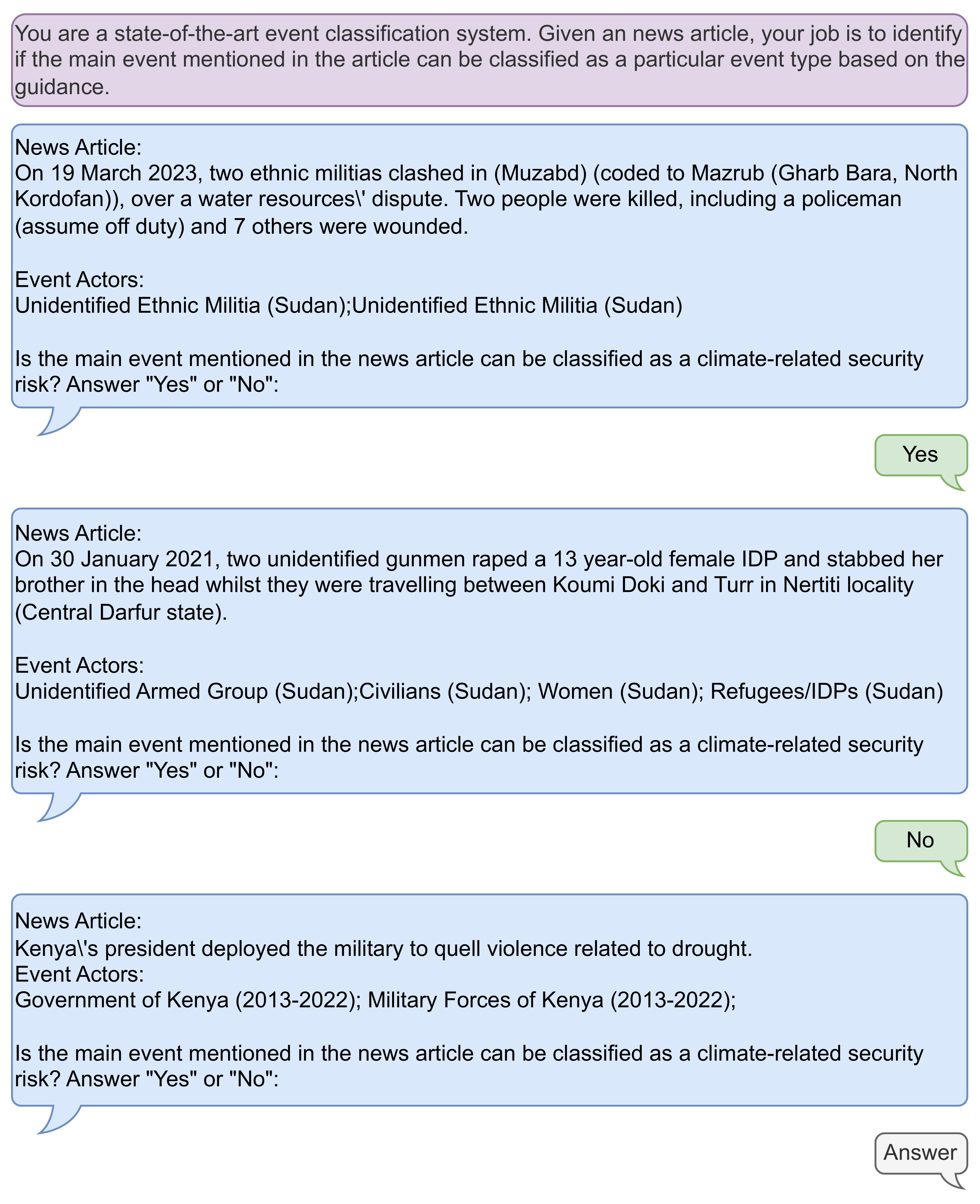}
    \caption{Our implementation of in-context learning in chat mode. Due to space constraints, we use an example with two-shots (one positive, and one negative example), and simplified prompts.}
    \label{fig:in_context}
\end{figure}

The implementation of In-context Learning can be found in Figure~\ref{fig:in_context}.
\label{apx:icl}
\subsection{Event-relevance Classification}
\textbf{System Prompt}

\noindent\small\fbox{
 \parbox{\linewidth}{
You are a state-of-the-art event detection system. Given a news article regarding a specific event, your job is to classify if the article is relevant based on the guidelines.

}
}
\clearpage
\subsubsection*{Zero-shot User Prompt}

\noindent\small\fbox{
 \parbox{\linewidth}{
Guidelines:\\
The article is relevant if:\\
1. the event it describes takes place in the Horn of Africa Region, which includes Djibouti, Eritrea, Ethiopia, Kenya, Somalia, Sudan, South Sudan, or Uganda.\\
2. the event it describes is violent and/or occurs in a conflict setting involving or aimed at a person or people (intended to intimidate/terrorize) instead of unassociated objects or things (general expression of anger, etc.).\\
3. the article describes a *specific event* and is not a summary of multiple events or different events, i.e., it is not describing multiple events or developments showing trends or general information. If an article mentions more than ONE event, it is not relevant in our setting.\\
\\
News Article:\\
\{DOCUMENT\}\\
\\
This event was POSSIBLY reported in \{COUNTRY\}.\\
\\
Is this article relevant based on the guidelines? Answer "Yes" or "No" in the following format (it must be valid XML):\\
<response>\\
<answer>Answer</answer>\\
<reason>reason for your selection</reason>\\
</response>

}
}

\subsubsection*{Six-shot User Prompt}
We adapt the zero-shot user prompt by remove the following sentence to create the six-shot user prompt, because the annotation for reasoning is not available.
\noindent\small\fbox{
 \parbox{\linewidth}{
<reason>reason for your selection</reason>
}
}

\subsubsection*{Six-shot Assistant Prompt}
\noindent\small\fbox{
 \parbox{\linewidth}{
<response>\\
<answer>\{ANSWER\}</answer>\\
</response>

}
}

\subsection{Event Type Classification}
\textbf{System Prompt}

\noindent\small\fbox{
 \parbox{\linewidth}{
You are a state-of-the-art event classification system. Given a news article, your job is to identify if the main event mentioned in the article can be classified as a particular event type based on the guidance.
}
}

\subsubsection*{Zero-shot User Prompt for Socio-political violence against women}
\noindent\small\fbox{
 \parbox{\linewidth}{
Guidance:\\
A Socio-political violence against women is civilian targeting event in which women and/or girls are the ‘target’ of the violence.\\
An event should be categorized as socio-political violence against women when:\\
- The victim(s) of the event are composed entirely of women/girls, or when the majority of victims are women/girls.\\
- The primary target was a woman/girl (e.g. a female politician attacked alongside two men working as bodyguards).\\
NOTE:\\
- DO NOT identify it as a socio-political violence against women event if the targeting of women or girls has the potential to be random. (Ex. woman killed while in a car that ran over an IED).\\
\\
News Article:\\
\{document\}\\
\\
Event Actors: \\
\{actor1\};\{actor2\}\\
\\
Is the main event mentioned in the news article can be classified as a socio-political violence against women? Answer "Yes" or "No" in the following format (it must be valid XML):\\
<response>\\
<event\_type>Answer</event\_type>\\
<reason>reason for your selection</reason>\\
</response>
}
}

\subsubsection*{Zero-shot User Prompt for climate-related security risk}
\noindent\small\fbox{
 \parbox{\linewidth}{
Guidance:\\
A climate-related security risk is a conflict event (like migration-induced conflict or pastoral conflict) influenced by environmental and climate-related factors.\\
These events should have two components: 1) a climate related phenomenon and 2) a conflict event, both of which are explicitly stated.\\
An event should be categorized as climate-related security risk when:\\
- A conflict event is influenced by environmental or climate related factors, which include, but are not limited to: drought, desertification, temperature rise, flooding.\\
- Conflict event centers around resources that have become limited due to environmental and climate-related factors, such as water access, grazing land, farmland, etc.\\
NOTE:\\
- DO NOT identify it as a climate-related security risk event if that is not directly influenced by climate-related factors, such as climate change protests.\\
\\
News Article:\\
\{document\}\\
\\
Event Actors: \\
\{actor1\};\{actor2\}\\
\\
Is the main event mentioned in the news article can be classified as a climate-related security risk? Answer "Yes" or "No" in the following format (it must be valid XML):\\
<response>\\
<event\_type>Answer</event\_type>\\
<reason>reason for your selection</reason>\\
</response>
}
}

\subsubsection*{Zero-shot User Prompt for religious conflict}
\noindent\small\fbox{
 \parbox{\linewidth}{
Guidance:\\
An event should be categorized as religious conflict as long as it meets any of the following requirements:\\
- Religion-related entity invlove in the conflict, which include, but are not limited to religious leaders, reglious military groups and religious staff; OR\\
- The conflict targets individuals who engage in religious practice or expressing their religious belief (e.g. pastor), no matter if the conflict itself is religiously motivated or not; OR\\
- It involves the enforcement of specific religious norms to force or prevent actions; OR\\
- The conflict happend at a religious institution.\\
NOTE:\\
- An event should be categoried as religious conflict when it meets any one of the above requirements.\\
- ALWAYS identity it as a religious conflict when military groups such as Al Shabaab and ISIS are involved.\\
- An event may also be categoried as a religous conflict  even though the conflict was not religiously motivated or targeted.\\
- DO NOT identify it as a religious conflict if it is explicitly mentioned in the article that the religious group / institution / person is a random target rather than a specific target. (Ex. mortar fire hits church in addition to many other nearby targets).\\
\\
News Article:\\
\{document\}\\
\\
Event Actors: \\
\{actor1\};\{actor2\}\\
\\
Is the main event mentioned in the news article can be classified as a religious conflict based on the guidance? Answer "Yes" or "No" in the following format (it must be valid XML):\\
<response>\\
<event\_type>Answer</event\_type>\\
<reason>reason for your selection</reason>\\
</response>
}
}

\subsubsection*{Zero-shot User Prompt for tribal/communal/ethnic conflict}
\noindent\small\fbox{
 \parbox{\linewidth}{
Guidance:\\
A tribal/communal/ethnic conflict is a dispute or violence involving ethnic, tribal, OR communal individuals/groups.\\
An event should be categorized as tribal/communal/ethnic conflict when:\\
- It falls into ANY of the following categories: tribal (including clans) OR communal OR ethnic.\\
NOTE:\\
- Disputes or violence can be one-sided from ethnic, tribal (including clans), OR communal individuals/groups.\\
- If the actor names are confirmed rather than presumed, please reference them to categorize a tribal/communal/ethnic conflict.\\
- DO NOT make conclusions based on presumed information.\\
\\
News Article:\\
\{document\}\\
\\
Event Actors: \\
\{actor1\};\{actor2\}\\
\\
Is the main event mentioned in the news article can be classified as a tribal/communal/ethnic conflict? Answer "Yes" or "No" in the following format (it must be valid XML):\\
<response>\\
<event\_type>Answer</event\_type>\\
<reason>reason for your selection</reason>\\
</response>
}
}

\subsubsection*{Six-shot User Prompt}
We remove the following sentence from the zero-shot user prompts to create the six-shot prompts for each event type, because the annotation for reasoning is not available.
\noindent\small\fbox{
 \parbox{\linewidth}{
<reason>reason for your selection</reason>
}
}

\subsubsection*{Six-shot Assistant Prompt}
\noindent\small\fbox{
 \parbox{\linewidth}{
<response>\\
<answer>\{ANSWER\}</answer>\\
</response>
}
}

\section{Modeling Details}
\subsection{BERT, Roberta}

We fine-tune BERT and RoBERTa models for both classification tasks. Due to the limited size of the training dataset, we restrict parameter updates to the last two layers. Early stopping is applied and model hyperparameters are chosen by optimizing the F1 score on the development set using grid-search. The final chosen hyperparameters are listed in Table~\ref{tab:hp_search_bert_roberta} and the model is trained on a single AWS p3.2xlarge machine, equipped
with a single NVIDIA V100 GPU with 16 GB of
GPU memory:

\begin{table}[!ht]
    \centering
    \renewcommand{\arraystretch}{1.1}
    \resizebox{0.45\textwidth}{!}{
    \begin{tabular}{|l|c|c|c|c|}
    \hline
    \multirow{2}{*}{} & \multicolumn{2}{c|}{\textbf{Relevance}} & \multicolumn{2}{c|}{\textbf{Event Type}} \\
    \cline{2-5}
    & \textbf{BERT} & \textbf{RoBERTa} & \textbf{BERT} & \textbf{RoBERTa} \\
    \hline
    Learning rate            & 0.001 & 0.001 & 0.001 & 0.001 \\
    Learning rate decay      & 0.05  & 0.05  & 0.05  & 0.05 \\
    Epoch                    & 100   & 100   & 100   & 100 \\
    Batch size               & 32    & 8     & 4     & 4 \\
    \hline
    \end{tabular}}
    \caption{Hyperparameters setting for Relevance and Event Type Classification for BERT and RoBERTa.}
    \label{tab:hp_search_bert_roberta}
\end{table}

\subsection{T5}
\label{sec:t5_questions}

\begin{table*}[!ht]
\resizebox{\textwidth}{!}{%
\begin{tabular}{|l|l|l|}
\hline
Task & Classes & Questions \\ \hline
Relevance Classification & Yes/No & Is the event relevant? \\ \hline
\multirow{4}{*}{Event Type Classification} & Tribal/Communal/Ethnic Conflict & Is the event Tribal/Communal/Ethnic Conflict? \\ \cline{2-3} 
 & Religious Conflict & Is the event Religious Conflict? \\ \cline{2-3} 
 & Socio-political Violence Against Women & Is the event Socio-political Violence Against Women? \\ \cline{2-3} 
 & Climate-Related Security Risks & Is the event Climate-Related Security Risks? \\ \hline
\end{tabular}%
}
\caption{Questions used for relevance and event-type classification tasks for T5.}
\label{tab:t5_questions}
\end{table*}

We format both relevance classification and event-type classification tasks as Question-Answering tasks for encoder-decoder models like T5. Table~\ref{tab:t5_questions} shows all the questions we prompt the T5 model during training and inference. We apply early stopping to select the best model checkpoint based on the best F1 score on the development set. The hyperparameters of the models are selected based on optimizing the F1 score on the development set via grid-search. Details of the selected hyperparameters are provided in Table~\ref{tab:hp_search_t5} and the model is trained on a single AWS p3.2xlarge machine, equipped
with a single NVIDIA V100 GPU with 16 GB of
GPU memory. %

\begin{table}[!ht]
    \centering
    \resizebox{0.5\textwidth}{!}{
    \begin{tabular}{|l|c|c|}
    \hline
    \textbf{Parameter} & \textbf{Relevance Classification} & \textbf{Event Type Classification} \\
    \hline
    Learning rate        & 0.0001 & 0.0001 \\
    Learning rate decay  & 0.05   & 0.05   \\
    Epoch                & 15     & 15     \\
    Batch size           & 8      & 8      \\
    \hline
    \end{tabular}}
    \caption{Hyperparameters setting for Relevance and Event Type Classification for T5.}
    \label{tab:hp_search_t5}
\end{table}